\documentclass[conference]{IEEEtran}
\IEEEoverridecommandlockouts
\usepackage{cite}
\usepackage{amsmath,amssymb,amsfonts}
\usepackage{algorithmic}
\usepackage[bookmarks=false]{hyperref}
\usepackage{graphicx}
\usepackage{textcomp}
\usepackage{xcolor}
\usepackage{microtype}
\def\BibTeX{{\rm B\kern-.05em{\sc i\kern-.025em b}\kern-.08em
    T\kern-.1667em\lower.7ex\hbox{E}\kern-.125emX}}
\begin{document}

\title{Using Restart Heuristics to Improve Agent Performance in Angry Birds}

\author{\IEEEauthorblockN{Tommy Liu \quad Jochen Renz \quad Peng Zhang}
\IEEEauthorblockA{\textit{Research School of Computer Science} \\
\textit{Australian National University}\\
Canberra, Australia \\
tommy.liu$\|$jochen.renz$\|$p.zhang@anu.edu.au}
\and
\IEEEauthorblockN{Matthew Stephenson}
\IEEEauthorblockA{\textit{Department of Data Science and Knowledge Engineering} \\
\textit{Maastricht University}\\
Maastricht, The Netherlands \\
matthew.stephenson@maastrichtuniversity.nl}
}

\IEEEpubid{\begin{minipage}{\textwidth}\ \\[12pt]
978-1-7281-1884-0/19/\$31.00 \copyright 2019 IEEE
\end{minipage}}

\maketitle

\begin{abstract}
Over the past few years the Angry Birds AI competition has been held in an attempt to develop intelligent agents that can successfully and efficiently solve levels for the video game Angry Birds. Many different agents and strategies have been developed to solve the complex and challenging physical reasoning problems associated with such a game. However none of these agents attempt one of the key strategies which humans employ to solve Angry Birds levels, which is restarting levels. Restarting is important in Angry Birds because sometimes the level is no longer solvable or some given shot made has little to no benefit towards the ultimate goal of the game. This paper proposes a framework and experimental evaluation for when to restart levels in Angry Birds. We demonstrate that restarting is a viable strategy to improve agent performance in many cases.
\end{abstract}


\section{Introduction}\label{introduction}
The problem of creating artificial agents which can interact with physical environments whilst achieving certain objectives is a key topic of research in Artificial Intelligence (AI). These agents will have to combine aspects such as computer vision, machine learning, reasoning, predicted motion and manipulation, and making decisions using incomplete information in order to achieve goals. One such endeavour is creating agents to solve the game of Angry Birds. 

In the Angry Birds AI (AIBIRDS) competition, participants develop AI agents to autonomously play the physics-based simulation game of Angry Birds, with the ultimate goal of beating the best human players \cite{jrenz:aibirds}. This competition promotes research into designing agents that can successfully interact with a physical environment, where knowledge about the environment is limited by perception and the exact consequences of the almost unlimited number of actions are unknown. Angry Birds provides a simple and controlled environment compared to the real world. The idea is that future artificial intelligence systems will need to overcome similar problems when attempting to interact with the real world \cite{jrenz:aibirds}. 

Previous agents in the AIBIRDS competition have implemented a variety of techniques such as Bayesian reinforcement learning \cite{Tziortziotis-et-al:angryBER}, qualitative reasoning \cite{walega-et-al:qualitative}, heuristics \cite{Borovicka-et-al:datalab}, internal simulations  \cite{PolceanuBuche:decision} or a combination of other agents\cite{Stephenson2017CreatingAH}.

However we can see that none of these current agents utilise the ability to restart levels while they are playing them, only doing so when they are forced to by failing the level. This is something that human players frequently do for the two reasons we observed in section \ref{introduction}. Restarting a level involves losing any progress made, returning the level to a state as if the agent had not interacted with it.

We know that the best agents of today fail to outperform reasonable human players \cite{aibirds:results} (benchmarks show that agents perform about 30\% worse on the `Poached Eggs' series of levels). We also know that most human players typically employ the restart strategy many times per level. We suggest employing heuristics as a guide to tell agents when to restart a given level. 

We observe two different types of situations in which humans restart levels. When the human has a plan in mind and the shot made is poor or sub-optimal compared to what was expected. Alternatively, a human may believe that the score cannot be improved any further. Learning and problem solving models incorporating human elements have made attempts at integrating heuristics for restarts \cite{langley-seth:ethps}. In \cite{langley-seth:ethps}, it is noted that heuristics can reduce the chance of incorrect choices being made but can seldom eliminate them completely. In this paper we develop heuristics that help reduce the chance of incorrect choices being made for Angry Birds. 

We employ qualitative spatial reasoning to develop a physical model of the Angry Birds world in order to infer consequences \cite{de:pr} of shots. In past agents, qualitative spatial reasoning has been used to analyze stability of levels in Angry Birds and to determine weak points to aim at \cite{zhang-renz:stability}. \cite{walega-et-al:qualitative} expands on this by introducing more complex qualitative rules and representations to determine the best shots to make.
These works apply common sense methods of operating on qualitative representations of objects \cite{cag:qsr}\cite{walega-et-al:qualitative} in order to determine what will happen if a given force occurs. Qualitative reasoning has also been applied to more general problem solving such as physical puzzle problems \cite{xge:hio}. 

We employ qualitative spatial reasoning to determine if a level is solvable or not using only a single shot. We only consider a single shot because predicting outcomes of shots that are accurate enough to deal with multiple shot sequences is still an unsolved problem due to the lack of a forward model. We build upon past work that determines the stability of a level along with qualitative reasoning rules for shots in Angry Birds. We also develop and test a set of heuristics that determines what shots may be considered sub-optimal.

We present and test three distinct heuristics in this paper, the most notable of which is the 'solvability' heuristic which  determines, using qualitative spatial reasoning, whether a level is likely solvable in one shot or not. This acts as a guide for restarting by telling us situations in which there is no reason to continue the level. It is obviously most useful when only one bird is left. The other two heuristics included in this module are based upon `common sense' reasoning rules on level states. These operate on state information such as score and number of objects in a level and informs us how much impact each action had.

An artificial agent must know in advance, or be capable of dealing with, the consequences of its actions. It must also be able to act upon and resolve unexpected or negative outcomes from the actions it performs. Our heuristics provide another measure for the outcomes of the actions that an agent performs in Angry Birds, allowing for more informed restarts to be made. 

The ideas presented here may also be relevant to more general AI situations. By considering restarts, we have to carefully consider what set of actions would be considered good and bad choices to make. We need to carefully evaluate if there is any benefit to trying alternative strategies to solve the given problem or to continue with our current strategy. For example agents playing other video games may want to consider restarting levels if certain objectives cannot be achieved any more. This is also relevant to agents that may wish to interact with the physical world. It is critical to take into account the consequences of any actions both before and after they are carried out in order to determine the next best steps.

In the rest of this paper, section \ref{background} provides background and further information regarding Angry Birds alongside the scope of this paper. Section \ref{methodology} describes the methodology we used for reasoning about structures and levels alongside our original contributions in the form of heuristics. Section \ref{testingandresults} tabulates the results of our testing. Sections \ref{discussionandfuturework} and \ref{conclusion} discuss our results and what it means for future works.

\section{Background}\label{background}

\begin{figure}
\begin{center}
\includegraphics[scale=0.52]{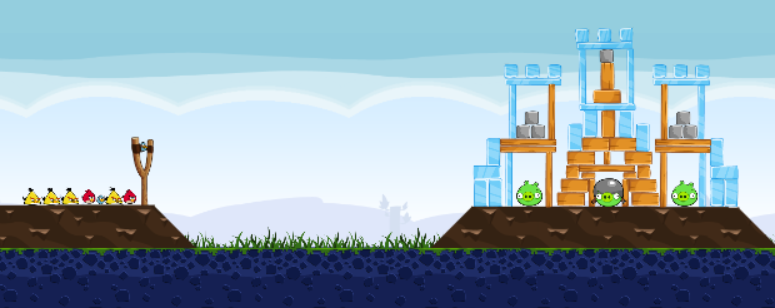}
\end{center}
\caption{An image of an Angry Birds level}
\end{figure}

\subsection{Angry Birds} is a physics based puzzle game where the player controls the angle and strength that a bird is released from a fixed position slingshot to shoot birds at levels consisting of structures and pigs (Fig. 1). The goal of the game is to kill all of the green pigs (targets) within each level given a limited number of birds (shots) whilst maximising the structure damage to achieve a high score. Blocks in Angry Birds (including pigs) are destroyed if they receive a sufficiently large enough impact or force acting on them, smaller forces can accumulate to eventually destroy large or tough blocks.
Scores are awarded at the end of each level based on how many birds have not been used and how much damage was inflicted on the level. There are also different types of blocks and birds which interact with each other in different ways. There are 3 main types of blocks in a level: wood, ice and stone, each with different properties such as toughness and vulnerabilities. The AIBIRDS competition has so far mainly focused on the following types of birds: red, blue, yellow, black and white. Each bird has a special power which is activated by tapping the screen
We define the performance of a player or agent as the average score they can obtain on given levels, along with how long on average it takes for them to solve levels or sets of levels.
Angry Birds is difficult because it involves an optimization problem (maximising score) which is hard to brute force because of the large number of possible shots and the time it takes for each attempt. This possibility space is further expanded by the many different types of blocks, birds and interactions between them. Furthermore, Angry Birds has been demonstrated to be an NP-Complete problem \cite{msteph:cc}. Agents will have to solve the problems presented in Angry Birds whilst outcomes are unclear in the environment, parameters are unknown and many steps in advance need to be considered, where each step corresponds to a single bird which can be used in the level \cite{jrez:abc}. 

\subsection{The AIBIRDS competition} \cite{aibirds:web} started out in 2012 and has run every year up to the present (2019). In this competition, participating agents are tasked with playing a number of unknown Angry Birds levels within a given time. They compete against each other based upon total points scored across all levels, typically 8 new levels per round. No internal information about these levels is provided and instead the AIBIRDS server provides screen-shots of the levels which the agents must parse, this means that agents get the exact same inputs that a human player would whilst playing the game. 
After a set amount of time has elapsed, the scores across levels are totalled and the agents are ranked based on these values, after several rounds where agents are eliminated, only one agent will remain. The goal of the AIBIRDS competition is to promote the development and exploration of agents and strategies which can deal with physical environments such as Angry Birds in general much like human actors/players \cite{jrenz:aibirds}. These techniques and ideas learned in the development of such agents are essential in the development of future AI systems as well \cite{jrenz:aibirds}.

\subsection{Current AIBIRDS Agents}
The currently most successful Angry Birds agents are heuristics based \cite{aibirds:web}. These heuristic agents work on a greedy basis regarding the current bird and make shot decisions using their heuristics without looking at future birds or outcomes. The issue with this is despite these strategies working with some number of levels and generally able to solve most competition levels, it is far from optimal. Angry Birds levels are usually designed with different optimal strategies in mind. For example on level 1-2 of the `Poached Eggs' (Fig. 2) series of levels it is possible to solve the level by shooting directly at the pigs. However the optimal strategy is to bounce a bird off the back wall which can kill all the pigs in one shot. 

\begin{figure}
\begin{center}
\includegraphics[scale=0.42]{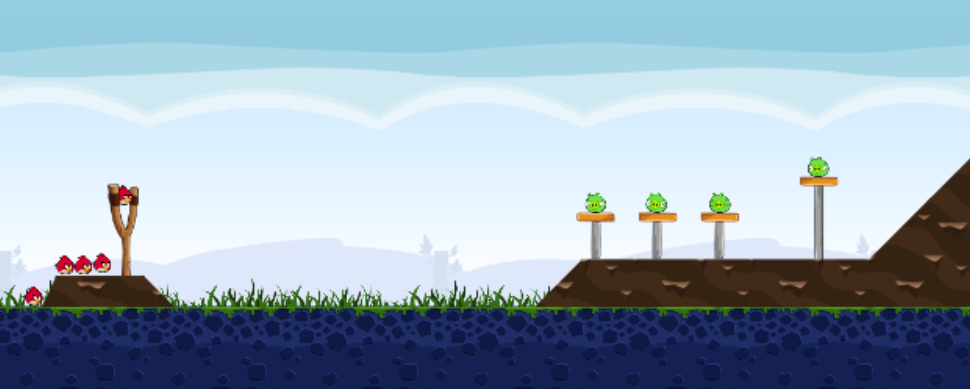}
\end{center}
\caption{Level 1-2 of the `Poached Eggs' level series}
\end{figure}

\subsection{Problem Exploration}
One core strategy that humans employ to find optimal strategies or possible strategies to levels is to explore possible solutions (shots) for the level. This can be likened to exploring the possible search space of the level, if shots don't have the desired impact or effect then a human would likely restart the level. We can liken human players to an agent with a knowledge base containing a set of heuristics and strategies along with some mechanism for learning new strategies by exploring different options on the level. The current AIBIRDS champion agent, Eagle's Wings, has a set of strategies which it decides between using some mechanism, however what it lacks is trying out new strategies and exploring different shots. We introduce a framework and a set of heuristics which encourage exploration of strategies by informing us when might be a good time to stop exploring a certain given strategy. This is especially important going forwards as current research has revealed what are known to be 'deceptive levels' where straightforward and traditional game playing strategies may fail to provide results \cite{msteph:dgames}. 

\section{Methodology}\label{methodology}

We first develop a module that predicts if a level is solvable in one shot, that is if using a single given bird whether we can destroy all the pigs in a level. This is useful in determining whether we should restart on the last bird of a level or not – this allows us to save some amount of time if we determine the level in unsolvable. It is difficult to determine whether a level is solvable in general because of the large search space and uncertainty of future states. We reason whether we can propagate sufficient forces to all the pigs in order to kill them, we build this off of the stability module in \cite{zhang-renz:stability} and the qualitative reasoning rules described in \cite{walega-et-al:qualitative}. Although it is possible to simulate the physics of Angry Birds precisely using the parameters of the Box2D physics model, these exact parameters are not known to the general public and agent developers. We employ qualitative reasoning methods to approximate solvability much quicker than simulations.

Next we develop a model of any given Angry Birds level by reading in all the Minimum Bounding Rectangles (MBR) blocks (that is wood, stone and ice) of objects. Our model develops matrices which tell us which blocks will be affected if another given block receives a force. We develop these matrices using the following relations:

\textbf{1. Direct Propagation:} We say if a block $o_1$ is in contact with a block $o_2$ and $o_1$ is to the left, above or below $o_2$ then if block $x$ receives a rightwards force, $o_2$ will receive the same force in the same direction reduced by some factor $c_1$. For two objects $o_1, o_2$ with points $(x_1,y_1),(x_2,y_2)$ and force $f_1$ on $o_1$, if the following Boolean predicate holds:
\begin{gather}
\begin{split}
\text{Direct\_Propagation}(o_1,o_2) \equiv  o_1 \neq o_2 \wedge x_2 < x_1 + k 
\\
\wedge (y_1 + \text{height}(o_1)) \text{ overlaps } (y_2 + \text{height}(o_2))
\end{split}
\end{gather}
Then $o_2$ will experience a force $f_2$:
\begin{gather}
f_2 = (f_1 / n)\cdot c
\end{gather}
Where $n$ is the number of blocks which we calculate $f_2$ for, and $c$ is some constant $<1$ which represents force lost. This intuitively means that the force experienced by $o_1$ will be distributed among all the blocks that it is in contact with (except those that receive no force).

\textbf{2. Falling Propagation:} if a block $o_2$ is in the falling arc (fall\_arc($o_1$)) of $o_1$, then if $o_1$ falls in a given direction, $o_2$ will receive some factor of the force received by $o_1$ depending on how far $o_1$ falls. If the following Boolean predicate holds:
\begin{align}
\begin{split}
\text{Falling\_Propagation}&(o_1, o_2) \equiv o_1 \neq o_2 \\
&\wedge \neg \text{Direct\_Propagation}(o_1,o_2) \\
&\wedge x_2 < x_1 + height(x_1) \\
&\wedge \text{intersects}(o_2, \text{fall\_arc($o_1$)})
\end{split}
\end{align}
Then $o_2$ will experience a force $f_2$:
\begin{align}
\begin{split}
f_2 = f_1\cdot c_1 \cdot sin\left(\pi \frac{d}{\text{height($o_1$)}}\right)
\end{split}
\end{align}
$d$ is the horizontal distance from $o_1$ to $o_2$, and $c_1$ is some constant $<1$ which represents force lost. Intuitively what this means is that closer blocks will receive a greater force, this however does not take into account any potential energy gains in force due to gravity.

\textbf{3. Structure Falling Propagation:} For two objects $o_1,o_2$ and the support structure of $o_1$: $supp(o_1):$, if $o_2$ intersects with fall\_arc(supp($o_1$). If the following boolean predicate holds:
\begin{align}
\begin{split}
\text{Structure}& \text{\_Falling\_Propagation}(o_1,o_2) \equiv o_1 \neq o_2 \\
&\wedge \text{intersects(fall\_arc(supp($o_1$)), $o_2$)} 
\end{split}
\end{align}
Then $o_2$ will experience $f_2 = f_1$. We do not do any complex calculations here because it is difficult to account for the many situations that this scenario occurs, since different structure types would result in different velocities at which they will fall down along with potential increases in energy due to gravity. 

\textbf{4. Thrown Blocks Propagation:} if a block $o_2$ is within a threshold vertical and horizontal distance of $o_1$ and $o_1$ is on the left of $o_2$ and $o_1$ is sufficiently small (of size less than $s_1$. Then, if $o_1$ receives a force, $o_1$ might be `thrown' a distance of $d$ and a height of $h$. This results in $o_2$ receiving a force in the same direction reduced by some factor. If the following Boolean predicate holds:
\begin{align}
\begin{split}
\text{Thrown\_blocks\_propagation}(o_1, o_2) \equiv o_1 \neq o_2 \\
\wedge \text{area}(o_1) < s_1 \wedge x_2 < x_1 + d \wedge y_2 < y_1 + h 
\end{split}
\end{align}
Then $o_2$ will experience $f_2$: 
\begin{align}
\begin{split}
f_2 = f1 \cdot c \cdot c_l \cdot d
\end{split}
\end{align}
Where $c_l$ is some constant of loss $<1$ for each unit the block travels through the air and $c$ is energy lost during the contact between $o_1$ and $o_2$.

\begin{figure}
\begin{center}
\includegraphics[scale=0.33]{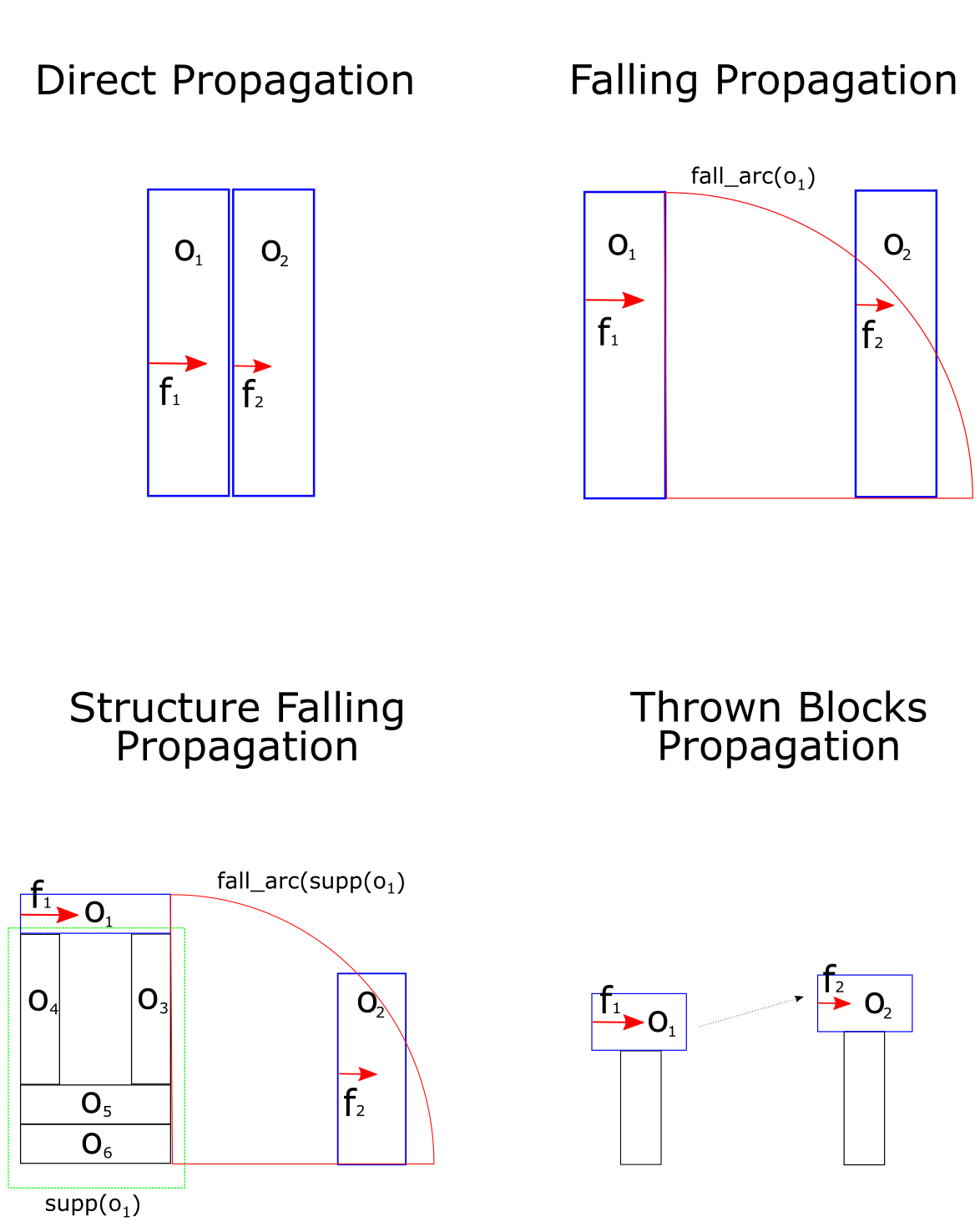}
\caption{Diagrams representing the propagation rules}
\end{center}
\end{figure}

These rules work alongside the previous rules for falling structures in \cite{zhang-renz:stability} to determine full propagation of forces through structures. We model the force propagated from a bird fired from a sling as a rightwards force originating from the point of impact. This force propagates using the above rules until the force is reduced to below some threshold. We say a level is solvable in one shot if all the pigs are killed (receive sufficient force to kill them), thus if a level is not solvable and we only have one shot left then we should restart. Using our rules we will tend to overestimate the solvability of levels, this is because many of these actions may not necessarily be possible (due to unknown factors such as friction) yet fall under our reasoning rules. For example we see in Fig. 2 the entire structure is not likely to fall, only parts of it. Overestimate when applying this to restarts is beneficial because the cost of restarting when the level is in fact solvable (time it takes to make all previous shots) plus any further time the agent spends on the level is much higher than not restarting when the level is unsolvable (time it takes to make one shot). Because of this, during much of our testing we used large placeholder values for $f_2$ close to 1 by setting the values for $c$ also to 1 to see the total effects of the forces propagating through the entire level.

Our algorithm currently does not take into account the following: TNT blocks, Black Birds, White Birds, Blue birds splitting apart and bird rebounds past the initial impact (i.e., birds bouncing off a wall to hit the right side of a block). However it is simply a matter of integrating and developing rules to take each of these into account to address our current limitations.

In order to further reduce our search space we determine which blocks are reachable from the sling. It is likely a waste of computational time to run our Force Propagation Algorithm on blocks that we cannot hit with a bird. We find these reachable blocks using a brute force search through all the blocks to see if there is a viable trajectory. We estimate trajectories using the trajectory module provided by the AIBIRDS framework. For each block we only use trajectories that hit the top center and left center points of every block. We only consider two points to reduce the total number of trajectories that we have to search through. This technique is an adapted version of the one used for reachable pigs in \cite{Stephenson2017CreatingAH}. We then check what blocks these trajectories intersect with, these are in theory all the blocks that can be reached. The problem with this is that the trajectory module is not perfect because we do not know the exact parameters of Angry Birds. Many agents implement their own trajectory module such as \cite{Borovicka-et-al:datalab} because of this limitation. We could address this problem by using any one of these better trajectory modules.

We implement all our models along with the algorithm to determine propagation of forces throughout a level in a single 'solvability' heuristic that is evaluated before a shot is made. We then develop rule-based heuristics that are evaluated after a shot is made. These post-shot heuristics are based on level states before and after the shot.

Finally we implement a framework to test the application of restarts on several agents: Seabirds, Datalab, Eagle's Wings and the Naive Agent. Eagle's Wings is the currently best performing agent having won the 2017 and 2018 AIBIRDS competitions. We test restarts with heuristics based on score changes, level differences and different types of actions performed.

\subsection{Heuristics}

We make decisions using our model and the qualitative rules described above before an agent makes a shot within a level. We also come up with rules which tell us if a we should restart a level after we make a shot. In the end we combine our heuristics together to form a single restart heuristic that tells us if we should restart a level at any given time. Our heuristics are described in more details below.

\subsubsection{Solvability Heuristic}

We apply a combination of our force propagation and reachability algorithms to determine solvability of a level. We say that a level is solvable if the forces resulting from hitting a reachable block propagate in such a way that all pigs receive some force that is sufficient to kill them. If we only have a single bird left, and this heuristic tells us that this level is not solvable then we should clearly restart the level. Conversely, because it is difficult to tell the future states of a level, we currently only run this heuristic if we have only one bird left.

\subsubsection{Score Heuristic} 
In addition to our pre-shot prediction-based solvability heuristic, we employ post-shot heuristics informing us when we should restart. One aspect of increasing performance is to increase total scores achieved by each agent. We calculate some threshold for each different color of bird and disallow shots which result in score increases less than our threshold. In theory, with learning agents, this heuristic will improve total scores in general because they will be forced to search out shots with higher scores. Higher score individual scores tend to have a very high correlation with higher overall scores . However, because this is a greedy approach these methods may suffer from the same problems current agents face and any deceptive levels mentioned in section \ref{background}. We attempt to mitigate this effect by introducing weighted decisions of when to restart in our Restart Heuristic later.

\subsubsection{Good Use Heuristic}

In Angry Birds every bird can usually be put to good use, with some exceptions \cite{Anderson-et-al:Deceptive} For example, it may be required to waste the first red bird of a level in order to solve the level using the second yellow bird. To solve an Angry Birds level, we must damage some structures and change the initial state of the level. We formulate some percentage change in a level where it is not likely a shot made was put to good use (e.g., the force of the bird only moved a single block a couple of pixels). We also say that a bird may not have been put to good use if we did not hit the block type it is good against (e.g., a blue bird hits a stone block). 
For score change $\Delta S$ and score threshold $T$, we give a value in the range $[0,1]$ with values closer to 0 being a 'worse' shot. This is given by the linear equation:
$$
\begin{aligned}
\text{ScoreH} = \frac{max((T - \Delta S), 0)}{T}
\end{aligned}
$$
\subsubsection{Restart Heuristic} 

We now combine our heuristics into a single heuristic in an attempt to make better predictions. We assign arbitrary weight to each of our heuristics to determine a total `restart score'. If this total score exceeds a pre-determined threshold we would give the signal to restart the level. We give this with \%Change being the percentage change in block locations and existence:
$$
\begin{aligned}
\text{Score}(\text{Screenshot}) = w_1 * (\text{\%Change}) + w_2 * (\text{Block Type Hit}) \\
+ w_3 * \text{ScoreH} + w_4 * \text{Solvable(Screenshot)}
\end{aligned}
$$

\section{Testing and Results}\label{testingandresults}
The agents we use for testing are:

\textbf{The Naive Agent:} A basic agent that is provided as part of the AIBIRDS framework that randomly determines which pig to shoot at.

\textbf{Seabirds Agent:} A middle-of-the-pack agent submitted to the AIBIRDS competition, it ranks at 15th out of 67 agents in the 2017 AIBIRDS benchmarks.

\textbf{Datalab Agent:} The winner of the AIBIRDS 2014 and 2015 competitions, uses a set of heuristics and a determination function to choose between them.

\textbf{Eagles Wings Agent:} The winner of the AIBIRDS 2017 and 2018 competitions, it is heavily based upon the Datalab agent and features a more complicated determination function to choose heuristics.

We use these agents as they provide a good measure of how our module works for different levels of agent performance. We use state of the art agents, Datalab and Eagles Wings as an upper limit to current AI performance. We use the Naive Agent as a benchmark as the worst performing agent since it is provided as part of the AIBIRDS framework (although there are worse agents). We use the Seabirds Agent to see its effects on a broader variety of strategies.

In order to test our solvability module, we ran tests with each agent using a test group and control group. Both groups use the exact same agents making the exact same shots, but the test group would restart when our module tells it to while the control group would execute the last shot without restarting. We do this because it is difficult to determine as an objective fact if a level is solvable or not without having tried all possible shots. We could manually classify if each level state is solvable however this is time consuming and humans also cannot determine this objective fact with certainty. Using this testing method we have the following outcomes:

\textbf{False Positive (FP)}: Module believes level not solvable, but the agent solves the level.

\textbf{False Negative (FN)}: Module believes level solvable, but the agent fails to solve the level.

\textbf{True Positive (TP)}: Module believes level not solvable, the agent fails to solve the level.

\textbf{True Negative (TN)}: Module believes level solvable, the agent solves the level.

The false positive class is what we are most concerned about. The cost of a false positive is the time taken to make all the shots up to that point, along with the rest of the time the agent may spend on the level. The cost of not restarting when the agent should have is just the time to run our module, since the agent would have made that shot anyway.

We ran the Naive and the Seabirds agent on the first 21 levels of the `Poached Eggs' series of levels. We do not test with Datalab and Eagle's Wings for this level set, because these two agents rarely need to use all the birds to solve the levels. We need to use harder levels to test these agents, hence we use the past AIBIRDS competition levels. We denote the number of trials - that is an iteration of a level being attempted by an agent as $n$.

\begin{table}[h!]
\caption{Percentage Results for Naive Agent using Solvability Module on `Poached Eggs levels' (n$\approx$1000).}\label{Naive}
\begin{center}
\begin{tabular}{|c|c|c|c|c|c|}
\hline
Level & TP            & TN            & FP            & FN            & TR   \\ \hline
1     & 0             & \textbf{1}    & 0             & 0             & N/A  \\ \hline
2     & 0             & \textbf{1}    & 0             & 0             & N/A  \\ \hline
3     & 0             & \textbf{1}    & 0             & 0             & N/A  \\ \hline
4     & 0             & \textbf{0.82} & 0.19          & 0             & 0.95 \\ \hline
5     & 0.30          & \textbf{0.4}  & 0.15          & 0.15          & 1.5  \\ \hline
6     & 0.05          & \textbf{0.7}  & 0.2           & 0.05          & 0.88 \\ \hline
7     & 0.30          & 0.10          & 0.052         & \textbf{0.55} & 2.4  \\ \hline
8     & \textbf{0.62} & 0.23          & 0.044         & 0.11          & 1.3  \\ \hline
9     & 0             & \textbf{0.79} & 0             & 0.21          & 1.3  \\ \hline
10    & \textbf{0.4}  & 0.21          & 0.26          & 0.13          & 1.5  \\ \hline
11    & 0.13          & \textbf{0.5}  & 0.25          & 0.12          & 1.0  \\ \hline
12    & 0             & \textbf{1}    & 0             & 0             & N/A  \\ \hline
13    & 0.16          & \textbf{0.39} & 0.065         & \textbf{0.39} & 1.4  \\ \hline
14    & \textbf{0.63} & 0.26          & 0.024         & 0.08          & 0.59 \\ \hline
15    & 0.10          & 0.16          & \textbf{0.71} & 0.03          & 1.0  \\ \hline
16    & 0.077         & \textbf{0.46} & \textbf{0.46} & 0             & 1.0  \\ \hline
17    & 0.035         & 0.36          & \textbf{0.52} & 0.086         & 0.67 \\ \hline
18    & 0.28          & 0.16          & \textbf{0.48} & 0.075         & 0.69 \\ \hline
19    & 0.069         & 0.10          & 0.069         & \textbf{0.76} & 1.2  \\ \hline
20    & \textbf{0.52} & 0.048         & 0.17          & 0.26          & 0.45 \\ \hline
21    & \textbf{0.56} & 0.077         & 0.13          & 0.24          & 2.3  \\ \hline
Average  & 0.20 & \textbf{0.47}       & 0.18          & 0.15          & 1.18  \\ \hline
\end{tabular}
\end{center}
\end{table}

The results for the Naive agent are shown in Table \ref{Naive}. We have a time ratio (TR) where we observed a difference in average time taken to solve a level between running the agent normally and running the agent and restarting with our solvability module. Ratio values above 1 mean that the agent with restarts was faster than without and values below 1 mean that the agent without restarts was faster. We have highlighted the majority class for each level in bold. The formula for time ratio is:
$$
\text{Time Ratio (TR)} = \frac{\text{Average time without restarts}}{\text{Average time with restarts}}
$$
With the Seabirds agent, we see a shift in the majority classes of outcomes in Table~\ref{CBirdsP}, notable we tend to see fewer False Positives and more True Negatives.
\begin{table}[!h]
\caption{Percentage Results for Seabirds agent using Solvability Module on `Poached Eggs levels' (n$\approx$850).}\label{CBirdsP}
\begin{center}
\begin{tabular}{|c|c|c|c|c|}
\hline
Level & TP            & TN             & FP            & FN            \\ \hline
1     & 0             & \textbf{1}     & 0             & 0             \\ \hline
2     & 0             & \textbf{0.83}  & 0.17          & 0             \\ \hline
3     & 0.2           & \textbf{0.8}   & 0             & 0             \\ \hline
4     & 0.14          & \textbf{0.57}  & 0             & 0.29          \\ \hline
5     & 0.29          & \textbf{0.29}  & 0             & 0.43          \\ \hline
6     & 0.36          & \textbf{0.45}  & 0.09          & 0.09          \\ \hline
7     & 0.2           & \textbf{0.3}   & 0.15          & 0.35          \\ \hline
8     & 0.16          & \textbf{0.58}  & 0.26          & 0             \\ \hline
9     & 0             & \textbf{0.67}  & 0             & 0.33          \\ \hline
10    & \textbf{0.34} & 0.14           & \textbf{0.34} & 0.17          \\ \hline
11    & 0             & \textbf{0.88}  & 0             & 0.12          \\ \hline
12    & 0.25          & \textbf{0.625} & 0             & 0.125         \\ \hline
13    & 0             & 0.31           & 0.19          & \textbf{0.5}  \\ \hline
14    & 0             & \textbf{1}     & 0             & 0             \\ \hline
15    & 0.13          & 0.22           & \textbf{0.65} & 0             \\ \hline
16    & 0.17          & \textbf{0.5}   & 0             & 0.33          \\ \hline
17    & 0             & \textbf{0.67}  & 0.13          & 0.2           \\ \hline
18    & 0.24          & 0.12           & \textbf{0.59} & 0             \\ \hline
19    & 0.25          & 0.062          & 0             & \textbf{0.69} \\ \hline
20    & \textbf{0.42} & 0.12           & 0.23          & 0.23          \\ \hline
21    & 0.27          & 0.27           & 0.08          & \textbf{0.38} \\ \hline
Average & 0.16 & \textbf{0.50}         & 0.14          & 0.20          \\ \hline
\end{tabular}
\end{center}
\end{table}

\begin{table}[!h]
\caption{Percentage Results for Seabirds agent using Solvability Module on 2017 AIBIRDS competition levels (n$\approx$300).}\label{CBirdscomp}
\begin{center}
\begin{tabular}{ccccc}
\hline
\multicolumn{5}{|c|}{2017 Grand Finals}                                                                                                                                          \\ \hline
\multicolumn{1}{|c|}{Level} & \multicolumn{1}{c|}{TP}            & \multicolumn{1}{c|}{TN}            & \multicolumn{1}{c|}{FP}            & \multicolumn{1}{c|}{FN}             \\ \hline
\multicolumn{1}{|c|}{1}     & \multicolumn{1}{c|}{0}             & \multicolumn{1}{c|}{\textbf{1}}    & \multicolumn{1}{c|}{0}             & \multicolumn{1}{c|}{0}              \\ \hline
\multicolumn{1}{|c|}{2}     & \multicolumn{1}{c|}{0}             & \multicolumn{1}{c|}{\textbf{0.69}} & \multicolumn{1}{c|}{0}             & \multicolumn{1}{c|}{0.31}           \\ \hline
\multicolumn{1}{|c|}{3}     & \multicolumn{1}{c|}{\textbf{0.97}} & \multicolumn{1}{c|}{0}             & \multicolumn{1}{c|}{0.03}          & \multicolumn{1}{c|}{0}              \\ \hline
\multicolumn{1}{|c|}{4}     & \multicolumn{1}{c|}{\textbf{0.45}} & \multicolumn{1}{c|}{0.05}          & \multicolumn{1}{c|}{\textbf{0.45}} & \multicolumn{1}{c|}{0.05}           \\ \hline
\multicolumn{1}{|c|}{5}     & \multicolumn{1}{c|}{\textbf{0.48}} & \multicolumn{1}{c|}{0.23}          & \multicolumn{1}{c|}{0.1}           & \multicolumn{1}{c|}{0.19}           \\ \hline
\multicolumn{1}{|c|}{6}     & \multicolumn{1}{c|}{\textbf{1}}    & \multicolumn{1}{c|}{0}             & \multicolumn{1}{c|}{0}             & \multicolumn{1}{c|}{0}              \\ \hline
\multicolumn{1}{|c|}{7}     & \multicolumn{1}{c|}{0.096}         & \multicolumn{1}{c|}{0.032}         & \multicolumn{1}{c|}{0}             & \multicolumn{1}{c|}{\textbf{0.87}}  \\ \hline
\multicolumn{1}{|c|}{8}     & \multicolumn{1}{c|}{0}             & \multicolumn{1}{c|}{\textbf{0.83}} & \multicolumn{1}{c|}{0.17}          & \multicolumn{1}{c|}{0}              \\ \hline
\multicolumn{1}{|c|}{Average}           & \multicolumn{1}{c|}{0.375} & \multicolumn{1}{c|}{\textbf{0.39}} & \multicolumn{1}{c|}{0.094} & \multicolumn{1}{c|}{0.18} \\ \hline
\multicolumn{1}{l}{}        & \multicolumn{1}{l}{}               & \multicolumn{1}{l}{}               & \multicolumn{1}{l}{}               & \multicolumn{1}{l}{}                \\ \hline

\multicolumn{5}{|c|}{2017 Semi Finals}                                                                                                                                           \\ \hline
\multicolumn{1}{|c|}{Level} & \multicolumn{1}{c|}{TP}            & \multicolumn{1}{c|}{TN}            & \multicolumn{1}{c|}{FP}            & \multicolumn{1}{c|}{FN}             \\ \hline
\multicolumn{1}{|c|}{1}     & \multicolumn{1}{c|}{0.083}         & \multicolumn{1}{c|}{0.29}          & \multicolumn{1}{c|}{0}             & \multicolumn{1}{c|}{\textbf{0.625}} \\ \hline
\multicolumn{1}{|c|}{2}     & \multicolumn{1}{c|}{0}             & \multicolumn{1}{c|}{\textbf{0.57}} & \multicolumn{1}{c|}{0.04}          & \multicolumn{1}{c|}{0.39}           \\ \hline
\multicolumn{1}{|c|}{3}     & \multicolumn{1}{c|}{0.25}          & \multicolumn{1}{c|}{0.03}          & \multicolumn{1}{c|}{0}             & \multicolumn{1}{c|}{\textbf{0.71}}  \\ \hline
\multicolumn{1}{|c|}{4}     & \multicolumn{1}{c|}{0}             & \multicolumn{1}{c|}{\textbf{0.84}} & \multicolumn{1}{c|}{0}             & \multicolumn{1}{c|}{0.16}           \\ \hline
\multicolumn{1}{|c|}{5}     & \multicolumn{1}{c|}{0.045}         & \multicolumn{1}{c|}{\textbf{0.59}} & \multicolumn{1}{c|}{0}             & \multicolumn{1}{c|}{0.36}           \\ \hline
\multicolumn{1}{|c|}{6}     & \multicolumn{1}{c|}{\textbf{0.85}} & \multicolumn{1}{c|}{0}             & \multicolumn{1}{c|}{0.059}         & \multicolumn{1}{c|}{0.088}          \\ \hline
\multicolumn{1}{|c|}{7}     & \multicolumn{1}{c|}{0.08}          & \multicolumn{1}{c|}{0}             & \multicolumn{1}{c|}{0}             & \multicolumn{1}{c|}{\textbf{0.92}}  \\ \hline
\multicolumn{1}{|c|}{8}     & \multicolumn{1}{c|}{0.17}          & \multicolumn{1}{c|}{\textbf{0.53}} & \multicolumn{1}{c|}{0.059}         & \multicolumn{1}{c|}{0.23}           \\ \hline
\multicolumn{1}{|c|}{Average}           & \multicolumn{1}{c|}{0.19} & \multicolumn{1}{c|}{0.36} & \multicolumn{1}{c|}{0.02} & \multicolumn{1}{c|}{\textbf{0.44}} \\ \hline
\end{tabular}
\end{center}
\end{table}

\begin{table}[!h]
\caption{Percentage Results the Eagles Wings agent using Solvability Module on 2017 AIBIRDS competition levels (n$\approx$300).}\label{Eaglewingscomp}
\begin{center}
\begin{tabular}{ccccc}
\hline
\multicolumn{5}{|c|}{2017 Grand Finals}                                                                                                                                         \\ \hline
\multicolumn{1}{|c|}{Level} & \multicolumn{1}{c|}{TP}            & \multicolumn{1}{c|}{TN}            & \multicolumn{1}{c|}{FP}            & \multicolumn{1}{c|}{FN}            \\ \hline
\multicolumn{1}{|c|}{1}     & \multicolumn{1}{c|}{0}             & \multicolumn{1}{c|}{0}             & \multicolumn{1}{c|}{\textbf{1}}    & \multicolumn{1}{c|}{0}             \\ \hline
\multicolumn{1}{|c|}{2}     & \multicolumn{1}{c|}{0}             & \multicolumn{1}{c|}{\textbf{1}}    & \multicolumn{1}{c|}{0}             & \multicolumn{1}{c|}{0}             \\ \hline
\multicolumn{1}{|c|}{3}     & \multicolumn{1}{c|}{0}             & \multicolumn{1}{c|}{\textbf{1}}    & \multicolumn{1}{c|}{0}             & \multicolumn{1}{c|}{0}             \\ \hline
\multicolumn{1}{|c|}{4}     & \multicolumn{1}{c|}{0.36}          & \multicolumn{1}{c|}{0}             & \multicolumn{1}{c|}{\textbf{0.45}} & \multicolumn{1}{c|}{0.18}          \\ \hline
\multicolumn{1}{|c|}{5}     & \multicolumn{1}{c|}{0.17}          & \multicolumn{1}{c|}{\textbf{0.5}}  & \multicolumn{1}{c|}{0.3}           & \multicolumn{1}{c|}{0}             \\ \hline
\multicolumn{1}{|c|}{6}     & \multicolumn{1}{c|}{\textbf{0.5}}  & \multicolumn{1}{c|}{0}             & \multicolumn{1}{c|}{\textbf{0.5}}  & \multicolumn{1}{c|}{0}             \\ \hline
\multicolumn{1}{|c|}{7}     & \multicolumn{1}{c|}{0.1}           & \multicolumn{1}{c|}{0.36}          & \multicolumn{1}{c|}{0}             & \multicolumn{1}{c|}{\textbf{0.55}} \\ \hline
\multicolumn{1}{|c|}{8}     & \multicolumn{1}{c|}{0}             & \multicolumn{1}{c|}{0}             & \multicolumn{1}{c|}{\textbf{1}}    & \multicolumn{1}{c|}{0}             \\ \hline
\multicolumn{1}{|c|}{Average}           & \multicolumn{1}{c|}{0.14} & \multicolumn{1}{c|}{0.36} & \multicolumn{1}{c|}{\textbf{0.41}} & \multicolumn{1}{c|}{0.091} \\ \hline
\multicolumn{1}{l}{}        & \multicolumn{1}{l}{}               & \multicolumn{1}{l}{}               & \multicolumn{1}{l}{}               & \multicolumn{1}{l}{}               \\ \hline
\multicolumn{5}{|c|}{2017 Semi Finals}                                                                                                                                          \\ \hline
\multicolumn{1}{|c|}{Level} & \multicolumn{1}{c|}{TP}            & \multicolumn{1}{c|}{TN}            & \multicolumn{1}{c|}{FP}            & \multicolumn{1}{c|}{FN}            \\ \hline
\multicolumn{1}{|c|}{1}     & \multicolumn{1}{c|}{\textbf{0.85}} & \multicolumn{1}{c|}{0}             & \multicolumn{1}{c|}{0.15}          & \multicolumn{1}{c|}{0}             \\ \hline
\multicolumn{1}{|c|}{2}     & \multicolumn{1}{c|}{0}             & \multicolumn{1}{c|}{0}             & \multicolumn{1}{c|}{0}             & \multicolumn{1}{c|}{\textbf{1}}    \\ \hline
\multicolumn{1}{|c|}{3}     & \multicolumn{1}{c|}{\textbf{1}}    & \multicolumn{1}{c|}{0}             & \multicolumn{1}{c|}{0}             & \multicolumn{1}{c|}{0}             \\ \hline
\multicolumn{1}{|c|}{4}     & \multicolumn{1}{c|}{0}             & \multicolumn{1}{c|}{\textbf{0.75}} & \multicolumn{1}{c|}{0}             & \multicolumn{1}{c|}{0.25}          \\ \hline
\multicolumn{1}{|c|}{5}     & \multicolumn{1}{c|}{0}             & \multicolumn{1}{c|}{\textbf{0.86}} & \multicolumn{1}{c|}{0}             & \multicolumn{1}{c|}{0.14}          \\ \hline
\multicolumn{1}{|c|}{6}     & \multicolumn{1}{c|}{0.23}          & \multicolumn{1}{c|}{0.14}          & \multicolumn{1}{c|}{\textbf{0.64}} & \multicolumn{1}{c|}{0}             \\ \hline
\multicolumn{1}{|c|}{7}     & \multicolumn{1}{c|}{\textbf{0.67}} & \multicolumn{1}{c|}{0}             & \multicolumn{1}{c|}{0.33}          & \multicolumn{1}{c|}{0}             \\ \hline
\multicolumn{1}{|c|}{8}     & \multicolumn{1}{c|}{0}             & \multicolumn{1}{c|}{\textbf{1}}    & \multicolumn{1}{c|}{0}             & \multicolumn{1}{c|}{0}             \\ \hline
\multicolumn{1}{|c|}{Average}           & \multicolumn{1}{c|}{\textbf{0.34}} & \multicolumn{1}{c|}{\textbf{0.34}} & \multicolumn{1}{c|}{0.14} & \multicolumn{1}{c|}{0.17} \\ \hline
\end{tabular}
\end{center}
\end{table}

We move onto testing our Restart Heuristic, we do this by comparing level
completion time and scores on various AIBIRDS competition level sets. We try various different combinations of weights and heuristics, the most notable increase in performance is noted in Table \ref{dlabres}. We denote the agent without our module as NR and with our module as R. Our value of $T$ varied based on the type of bird previously fired, the values used for red, blue, yellow and black birds are 5000, 6000, 7000, 10000 respectively.  We used the weights 0.2, 0.2, 0.2 and 0.4 for $w_1...w_4$ respectively.

Conversely we also note that is is possible to have a negative impact on scores and times by using our module. We tabulate the worst case performance change by using our module below in Table \ref{dlabresbad}.

\begin{table}[!h]
\caption{Score and Times for Datalab agent on AIBIRDS 2015 Grand Finals levels using score and good use heuristics (n$\approx$300). Score in points, Times in seconds.}\label{dlabres}
\begin{center}
\begin{tabular}{ccccc}
\hline
\multicolumn{5}{|c|}{2015 Grand Finals}                                                                                                               \\ \hline
\multicolumn{1}{|c|}{Level} & \multicolumn{1}{c|}{Score NR} & \multicolumn{1}{c|}{STD NR} & \multicolumn{1}{c|}{Score R} & \multicolumn{1}{c|}{STD R} \\ \hline
\multicolumn{1}{|c|}{1}     & \multicolumn{1}{c|}{77432}    & \multicolumn{1}{c|}{6534}   & \multicolumn{1}{c|}{78640}   & \multicolumn{1}{c|}{370}   \\ \hline
\multicolumn{1}{|c|}{2}     & \multicolumn{1}{c|}{78000}    & \multicolumn{1}{c|}{10148}  & \multicolumn{1}{c|}{78274}   & \multicolumn{1}{c|}{9739}  \\ \hline
\multicolumn{1}{|c|}{3}     & \multicolumn{1}{c|}{81569}    & \multicolumn{1}{c|}{6742}   & \multicolumn{1}{c|}{83482}   & \multicolumn{1}{c|}{7808}  \\ \hline
\multicolumn{1}{|c|}{4}     & \multicolumn{1}{c|}{N/A}      & \multicolumn{1}{c|}{N/A}    & \multicolumn{1}{c|}{N/A}     & \multicolumn{1}{c|}{N/A}   \\ \hline
\multicolumn{1}{|c|}{5}     & \multicolumn{1}{c|}{51765}    & \multicolumn{1}{c|}{3956}   & \multicolumn{1}{c|}{50364}   & \multicolumn{1}{c|}{4132}  \\ \hline
\multicolumn{1}{|c|}{6}     & \multicolumn{1}{c|}{39223}    & \multicolumn{1}{c|}{1118}   & \multicolumn{1}{c|}{46290}   & \multicolumn{1}{c|}{5727}  \\ \hline
\multicolumn{1}{|c|}{7}     & \multicolumn{1}{c|}{95269}    & \multicolumn{1}{c|}{7489}   & \multicolumn{1}{c|}{94056}   & \multicolumn{1}{c|}{4211}  \\ \hline
\multicolumn{1}{|c|}{8}     & \multicolumn{1}{c|}{34090}    & \multicolumn{1}{c|}{1066}   & \multicolumn{1}{c|}{36785}   & \multicolumn{1}{c|}{4211}  \\ \hline
\multicolumn{1}{|c|}{Total} & \multicolumn{1}{c|}{69910}    & \multicolumn{1}{c|}{22083}  & \multicolumn{1}{c|}{73636}   & \multicolumn{1}{c|}{19447} \\ \hline
\multicolumn{1}{l}{}        & \multicolumn{1}{l}{}          & \multicolumn{1}{l}{}        & \multicolumn{1}{l}{}         & \multicolumn{1}{l}{}       \\ \hline
\multicolumn{1}{|c|}{Level} & \multicolumn{1}{c|}{Time NR}  & \multicolumn{1}{c|}{STD NR} & \multicolumn{1}{c|}{Time R}  & \multicolumn{1}{c|}{STD R} \\ \hline
\multicolumn{1}{|c|}{1}     & \multicolumn{1}{c|}{147}      & \multicolumn{1}{c|}{151}    & \multicolumn{1}{c|}{136}     & \multicolumn{1}{c|}{97}    \\ \hline
\multicolumn{1}{|c|}{2}     & \multicolumn{1}{c|}{261}      & \multicolumn{1}{c|}{94.1}   & \multicolumn{1}{c|}{236}     & \multicolumn{1}{c|}{76.3}  \\ \hline
\multicolumn{1}{|c|}{3}     & \multicolumn{1}{c|}{249}      & \multicolumn{1}{c|}{139}    & \multicolumn{1}{c|}{321}     & \multicolumn{1}{c|}{157}   \\ \hline
\multicolumn{1}{|c|}{4}     & \multicolumn{1}{c|}{N/A}      & \multicolumn{1}{c|}{N/A}    & \multicolumn{1}{c|}{N/A}     & \multicolumn{1}{c|}{N/A}   \\ \hline
\multicolumn{1}{|c|}{5}     & \multicolumn{1}{c|}{384}      & \multicolumn{1}{c|}{171}    & \multicolumn{1}{c|}{498}     & \multicolumn{1}{c|}{149}   \\ \hline
\multicolumn{1}{|c|}{6}     & \multicolumn{1}{c|}{143}      & \multicolumn{1}{c|}{124}    & \multicolumn{1}{c|}{139}     & \multicolumn{1}{c|}{126}   \\ \hline
\multicolumn{1}{|c|}{7}     & \multicolumn{1}{c|}{136}      & \multicolumn{1}{c|}{63.2}   & \multicolumn{1}{c|}{115}     & \multicolumn{1}{c|}{23.8}  \\ \hline
\multicolumn{1}{|c|}{8}     & \multicolumn{1}{c|}{152}      & \multicolumn{1}{c|}{168}    & \multicolumn{1}{c|}{82.5}    & \multicolumn{1}{c|}{21.4}  \\ \hline
\multicolumn{1}{|c|}{Total} & \multicolumn{1}{c|}{1472}     & \multicolumn{1}{c|}{1069}   & \multicolumn{1}{c|}{1681}    & \multicolumn{1}{c|}{650}   \\ \hline
\end{tabular}
\end{center}
\end{table}

\begin{table}[!h]
\caption{Score and Times for Datalab agent on AIBIRDS 2014 Grand Finals levels using score and good use heuristics (n$\approx$300). Score in points, Times in seconds.}\label{dlabresbad}
\begin{center}
\begin{tabular}{ccccc}
\hline
\multicolumn{5}{|c|}{2014 Grand Finals}                                                                                                               \\ \hline
\multicolumn{1}{|c|}{Level} & \multicolumn{1}{c|}{Score NR} & \multicolumn{1}{c|}{STD NR} & \multicolumn{1}{c|}{Score R} & \multicolumn{1}{c|}{STD R} \\ \hline
\multicolumn{1}{|c|}{1}     & \multicolumn{1}{c|}{59999}    & \multicolumn{1}{c|}{7451}   & \multicolumn{1}{c|}{60542}   & \multicolumn{1}{c|}{15772} \\ \hline
\multicolumn{1}{|c|}{2}     & \multicolumn{1}{c|}{56370}    & \multicolumn{1}{c|}{7930}   & \multicolumn{1}{c|}{53757}   & \multicolumn{1}{c|}{8253}  \\ \hline
\multicolumn{1}{|c|}{3}     & \multicolumn{1}{c|}{64757}    & \multicolumn{1}{c|}{4742}   & \multicolumn{1}{c|}{64111}   & \multicolumn{1}{c|}{2058}  \\ \hline
\multicolumn{1}{|c|}{4}     & \multicolumn{1}{c|}{66620}    & \multicolumn{1}{c|}{N/A}    & \multicolumn{1}{c|}{N/A}     & \multicolumn{1}{c|}{N/A}   \\ \hline
\multicolumn{1}{|c|}{5}     & \multicolumn{1}{c|}{42922}    & \multicolumn{1}{c|}{7874}   & \multicolumn{1}{c|}{N/A}     & \multicolumn{1}{c|}{N/A}   \\ \hline
\multicolumn{1}{|c|}{6}     & \multicolumn{1}{c|}{67888}    & \multicolumn{1}{c|}{4714}   & \multicolumn{1}{c|}{66750}   & \multicolumn{1}{c|}{3434}  \\ \hline
\multicolumn{1}{|c|}{7}     & \multicolumn{1}{c|}{N/A}      & \multicolumn{1}{c|}{N/A}    & \multicolumn{1}{c|}{N/A}     & \multicolumn{1}{c|}{N/A}   \\ \hline
\multicolumn{1}{|c|}{8}     & \multicolumn{1}{c|}{70219}    & \multicolumn{1}{c|}{N/A}    & \multicolumn{1}{c|}{61570}   & \multicolumn{1}{c|}{N/A}   \\ \hline
\multicolumn{1}{|c|}{Total} & \multicolumn{1}{c|}{428775}   & \multicolumn{1}{c|}{32711}  & \multicolumn{1}{c|}{306630}  & \multicolumn{1}{c|}{29517} \\ \hline
\multicolumn{1}{l}{}        & \multicolumn{1}{l}{}          & \multicolumn{1}{l}{}        & \multicolumn{1}{l}{}         & \multicolumn{1}{l}{}       \\ \hline
\multicolumn{1}{|c|}{Level} & \multicolumn{1}{c|}{Time NR}  & \multicolumn{1}{c|}{STD NR} & \multicolumn{1}{c|}{Time R}  & \multicolumn{1}{c|}{STD R} \\ \hline
\multicolumn{1}{|c|}{1}     & \multicolumn{1}{c|}{170}      & \multicolumn{1}{c|}{222}    & \multicolumn{1}{c|}{323}     & \multicolumn{1}{c|}{348}   \\ \hline
\multicolumn{1}{|c|}{2}     & \multicolumn{1}{c|}{91.3}     & \multicolumn{1}{c|}{173}    & \multicolumn{1}{c|}{204}     & \multicolumn{1}{c|}{307}   \\ \hline
\multicolumn{1}{|c|}{3}     & \multicolumn{1}{c|}{267}      & \multicolumn{1}{c|}{334}    & \multicolumn{1}{c|}{151}     & \multicolumn{1}{c|}{78.0}  \\ \hline
\multicolumn{1}{|c|}{4}     & \multicolumn{1}{c|}{961}      & \multicolumn{1}{c|}{N/A}    & \multicolumn{1}{c|}{N/A}     & \multicolumn{1}{c|}{N/A}   \\ \hline
\multicolumn{1}{|c|}{5}     & \multicolumn{1}{c|}{805}      & \multicolumn{1}{c|}{368}    & \multicolumn{1}{c|}{N/A}     & \multicolumn{1}{c|}{N/A}   \\ \hline
\multicolumn{1}{|c|}{6}     & \multicolumn{1}{c|}{534}      & \multicolumn{1}{c|}{361}    & \multicolumn{1}{c|}{513}     & \multicolumn{1}{c|}{666}   \\ \hline
\multicolumn{1}{|c|}{7}     & \multicolumn{1}{c|}{N/A}      & \multicolumn{1}{c|}{N/A}    & \multicolumn{1}{c|}{N/A}     & \multicolumn{1}{c|}{N/A}   \\ \hline
\multicolumn{1}{|c|}{8}     & \multicolumn{1}{c|}{818}      & \multicolumn{1}{c|}{N/A}    & \multicolumn{1}{c|}{1014}    & \multicolumn{1}{c|}{N/A}   \\ \hline
\end{tabular}
\end{center}
\end{table}

\section{Performance and Analysis}\label{performanceandanalysis}
We see that in the cases of simple levels such as the `Poached Eggs` series Table \ref{Naive}, our module works well when restarting with exceptions of levels 14,17,18 and 20. These exceptions can be explained by current shortcomings in our solvability module, computer vision module and reachability algorithms. We note that as the number of false positives increase we tend to see a reduction in agent performance however this is not always the case. We see on level 20, despite the majority case being True Positives, we see a reduction in agent performance. It may be the case that the false positives are more impactful on these levels than others because of a more complex level structure. Nonetheless we note that any proportion of False Positives tend to have a negative impact on performance. We also note the relatively low total number of false positives compared to the others which results in possible better performance for the agent. We see that on average the time ratio is greater than 1 (1.18) across the 17 levels that we have utilised it in. This means that using our solvability module we tend to take less time to solve levels than without.

Moving from the Naive Agent to the Seabirds agent in Table \ref{CBirdsP}, we see a tendency for fewer False Positives and more true negatives. Because we know that the Seabirds agent performs better than the Naive Agent, this perhaps indicates that for better agents, our solvability algorithm tends to be less inaccurate. This also illustrates differences in ground truths between different types of agents because some situations are more solvable than others for specific strategies. 

On more difficult level sets such as the competition levels, it is harder to attribute with certainty the efficacy or if the effects are due to random chance. Mentioned in section III our module does not implement several features of the Angry Birds game, the competition levels tend to utilise a great deal of these features. Nonetheless we observe that the number of false positives on competition levels is low with the exception of the 2017 Grand Finals in Table \ref{Eaglewingscomp}. Using Seabirds we see in Table \ref{CBirdscomp} that the FP class is nearly 0 in most cases.

Tables \ref{dlabres} and \ref{dlabresbad} show that our restart heuristic can have both positive and negative effects on the performance of the agents. We have observed that the agent with our restart module fails to complete several levels, resulting in an N/A value for scores. This was mostly caused by constantly restarting and making the same shots during the playing of a level - we imposed a time threshold where if the level was not completed, the agent would move on. This suggests that our heuristics may not be a good fit for the current generation of agents. A major factor contributing to this may be that the agents we are testing on are not 'aware' that they are being restarted because our heuristics have not been fully integrated into the core logic of the agents. As a result of this, sometimes the exact same strategies/shots would be made as observed in some occasions. Conversely we see an increase in level scores in Table \ref{dlabres}, this increase may also have caused an increase in mean times taken. We may explain this by the agent spending more time searching for the best shots to make because of the restarts.

\section{Discussion and Future Work}\label{discussionandfuturework}
As we have demonstrated, restarting when the level is no longer possible could result in an increase in the performance of an agent. Logically speaking, if the level is not possible then we can equate this to a lost level and restarting would clearly be the optimal choice. However deciding when a level is unsolvable is a difficult task, we have demonstrated a method that approximates this problem to some degree of accuracy. With more work on extending our solvability module, along with tuning weights on each individual heuristics, we believe that we can achieve a high degree of accuracy on all Angry Birds Levels.

We believe that in order to improve further, Angry Birds agents will have to take what we have presented in this paper into account. As long as humans employ this strategy and artificial agents do not, it may not be possible to surpass them. In order to learn from restarts, agent logic and restart logic must be fully coupled. In our testing with the agent separated from the restart module, agents would not integrate the knowledge gained from restarting into future attempts. This results in less than optimal and sometimes even negative effects on agent performance. This is a problem that can be solved in the development of future agents by integrating our restart heuristics into the learning models of the agent.

AIBIRDS evaluates the performance of an agent purely on the total sum of scores achieved in the time limit allowed. It may be the case that restarting actually has an overall negative effect during the time limit allowed. For example restarting failed level attempts constitutes more time invested in a single level, resulting in less time invested in levels in which the agent might actually perform very well in. 
Given sufficient time to solve a level there is no need to restart because eventually an optimal solution will be found however it is rarely the case that time is infinite. On many levels we will observe humans attempt a level many times by restarting in order to achieve series of what the human would consider to be optimal shots to get the overall highest score. This is a key characteristic of humans and perhaps a defining characteristic of intelligence in order to recognize the best use of time invested to achieve a goal.

\section{Conclusion}\label{conclusion}

This paper has presented an initial framework which determines when may be a good time to restart the level using the solvability of a level and other simple heuristics. This knowledge can be used to reduce the amount of time wasted by continuing to play levels which are unsolvable, or search for better alternatives to shots. 

We will continue developing our solvability algorithm to work with all Angry Birds levels. The causes of many negative results can be identified, this means that we know what we need to work on to improve and expand our algorithm in the future. We hope that future agents implement and use our module as core components in the pursuit of agents that perform better than humans.

\end{document}